\documentclass[lettersize,journal]{IEEEtran}
\usepackage{cite}
\usepackage{amsmath,amssymb,amsfonts}
\usepackage{algorithm}
\usepackage{algorithmic}
\usepackage{graphicx}
\usepackage{subfigure}
\usepackage{textcomp}
\usepackage{xcolor}
\usepackage{booktabs}

\def\BibTeX{{\rm B\kern-.05em{\sc i\kern-.025em b}\kern-.08em
    T\kern-.1667em\lower.7ex\hbox{E}\kern-.125emX}}

\begin{document}

\title{Knowledge Enhanced Semantic Communication Receiver}

\author{Bingyan Wang, Rongpeng Li, Jianhang Zhu, Zhifeng Zhao, and Honggang Zhang
\thanks{B. Wang, R. Li and J. Zhu are with the College of Information Science and Electronic Engineering, Zhejiang University, Hangzhou 310027, China (e-mail: \mbox{\{wangbingyan, lirongpeng, zhujh20\}@zju.edu.cn}).}
\thanks{Z. Zhao and H. Zhang are with Zhejiang Lab, Hangzhou, China as well as the College of Information Science and Electronic Engineering, Zhejiang University, Hangzhou 310027, China (e-mail: \mbox{zhaozf@zhejianglab.com}, \mbox{honggangzhang@zju.edu.cn}).}}

\maketitle

\begin{abstract}
In recent years, with the rapid development of deep learning and natural language processing technologies, semantic communication has become a topic of great interest in the field of communication. Although existing deep learning-based semantic communication approaches have shown many advantages, they still do not make sufficient use of prior knowledge. Moreover, most existing semantic communication methods focus on the semantic encoding at the transmitter side, while we believe that the semantic decoding capability of the receiver should also be concerned. In this paper, we propose a knowledge enhanced semantic communication framework in which the receiver can more actively utilize the facts in the knowledge base for semantic reasoning and decoding, on the basis of only affecting the parameters rather than the structure of the neural networks at the transmitter side. Specifically, we design a transformer-based knowledge extractor to find relevant factual triples for the received noisy signal. Extensive simulation results on the WebNLG dataset demonstrate that the proposed receiver yields superior performance on top of the knowledge graph enhanced decoding. 
\end{abstract}


\section{Introduction}

Benefiting from the rapid development of deep learning (DL) and natural language processing (NLP), semantic communications emerge with a special emphasis on the successful delivery of the semantics of a message, rather than the conventional bit-level accuracy in traditional communication. There have been some interesting studies on semantic communication\cite{xie2021deep,zhou2021semantic,jiang2022deep,zhou2022adaptive,hu2022robust}. Among them, one of the popular paradigms belongs to the DL-based joint source-channel coding (JSCC). For example, Ref.~\cite{xie2021deep} proposes a transformer-based semantic communication system for text transmission. Ref.~\cite{zhou2021semantic} introduces a semantic communication system based on Universal Transformer (UT) with an adaptive circulation mechanism. In order to reduce the semantic transmission error, Ref.~\cite{jiang2022deep} exploits hybrid automatic repeat request (HARQ), while Ref.~\cite{zhou2022adaptive} introduces an adaptive bit rate control mechanism. Moreover, Ref.~\cite{hu2022robust} proposes a masked autoencoder (MAE) based system to robustly combat the possible noise.  Notably, a key assumption of these studies lies in that both transmitter and receiver share common knowledge. On top of this assumption, the existing semantic communication methods jointly train the DL-based transmitter and receiver, and have proven their superiority over traditional communication methods. However, the receiver is still lacking the comprehensive knowledge understanding and reasoning ability, and cannot make full use of the implicit prior knowledge in complex sentences. 

In order to improve the capability of knowledge understanding and reasoning, some studies propose to introduce the knowledge graph (KG), which stores human knowledge with a graph structure composed of entities and relationships \cite{ji2021survey}, into semantic communication. In KGs, each fact is abstracted into a triple in the form of (entity-relationship-entity). For example, Ref.~\cite{wang2021performance} utilizes knowledge triples to represent the semantic information and evaluates the importance of each triple by an attention policy gradient algorithm. Ref.~\cite{zhou2022cognitive} proposes a semantic communication framework by encoding texts into KGs. Ref.~\cite{liang2022life} introduces a knowledge reasoning based semantic communication system. In Ref.~\cite{jiang2022reliable}, a reliable semantic communication system based on KG is proposed, which can adaptively adjust the transmitted triples according to channel quality. In Ref.~\cite{choi2022unified}, the authors exploit the knowledge base by leveraging a logic programming language. In Ref.~\cite{muppavarapu2021knowledge}, the authors propose a semantic similarity-based approach to automatically identify and extract the most common concepts from the knowledge base. 

Knowledge graphs have somewhat improved the capability of semantic communication systems to handle common knowledge. However, most existing works only consider optimizing the transmitter while ignoring the receiver. Typically, their transmitters achieve the semantic encoding by capturing and embedding the factual triples from the sentences with knowledge graphs. Nonetheless, it is a great challenge for a knowledge base to cover all the semantic information of a sentence, and the information missing may be detrimental to the communication efficiency. For example, a sentence like “She loves him” can't be represented by any factual triples in the knowledge base, but it might be also the vital element in a transmitted text, leading the failure unacceptable. Instead, knowledge graphs can only describe the semantics in those simple declarative sentences. Therefore, sending messages that are only encoded by knowledge graph-based triples may cause extra semantic loss.

Therefore, in order to address these issues, we propose a novel receiver-side scheme for semantic communication based on KG. Different from existing works that extract factual triples from the transmitter side as the semantic representations, we apply a knowledge extraction module at the receiver side as a semantic decoding assistant to avoid the injection of extra semantic noise and enhance the model's robustness in low SNR environment, making semantic communications more effective. By doing so, the knowledge in the knowledge base can be utilized for decoding on the basis of only affecting the parameters rather than the structure of the deep neural networks (DNN) at the transmitter side. Unlike the transmitter-side schemes\cite{zhou2022cognitive, jiang2022reliable, liang2022life}, as the received content is inevitably polluted by noise, it remains essential to accurately extract the factual triples from noisy content before leveraging them to complement the decoding procedure. Therefore, rather than focusing on each word in a sentence, it is better to consider extracting the semantic knowledge representation of the whole sentence from a novel perspective. For this purpose, we utilize transformer encoders to get the implicit semantic representation of a sentence. By integrating KG and knowledge extractor into the conventional semantic decoder, the receiver can extract knowledge from noisy messages and enhance the decoding capability.

The remainder of the paper is organized as follows. The system model and problem formulation are given in Section II. Section III describes the DNN structure of a knowledge enhanced semantic receiver. Section IV discusses the simulation settings and experimental results. Section V concludes the paper.

\section{System Model and Problem Formulation}

\begin{figure}[t]
\centerline{\includegraphics[width=\columnwidth]{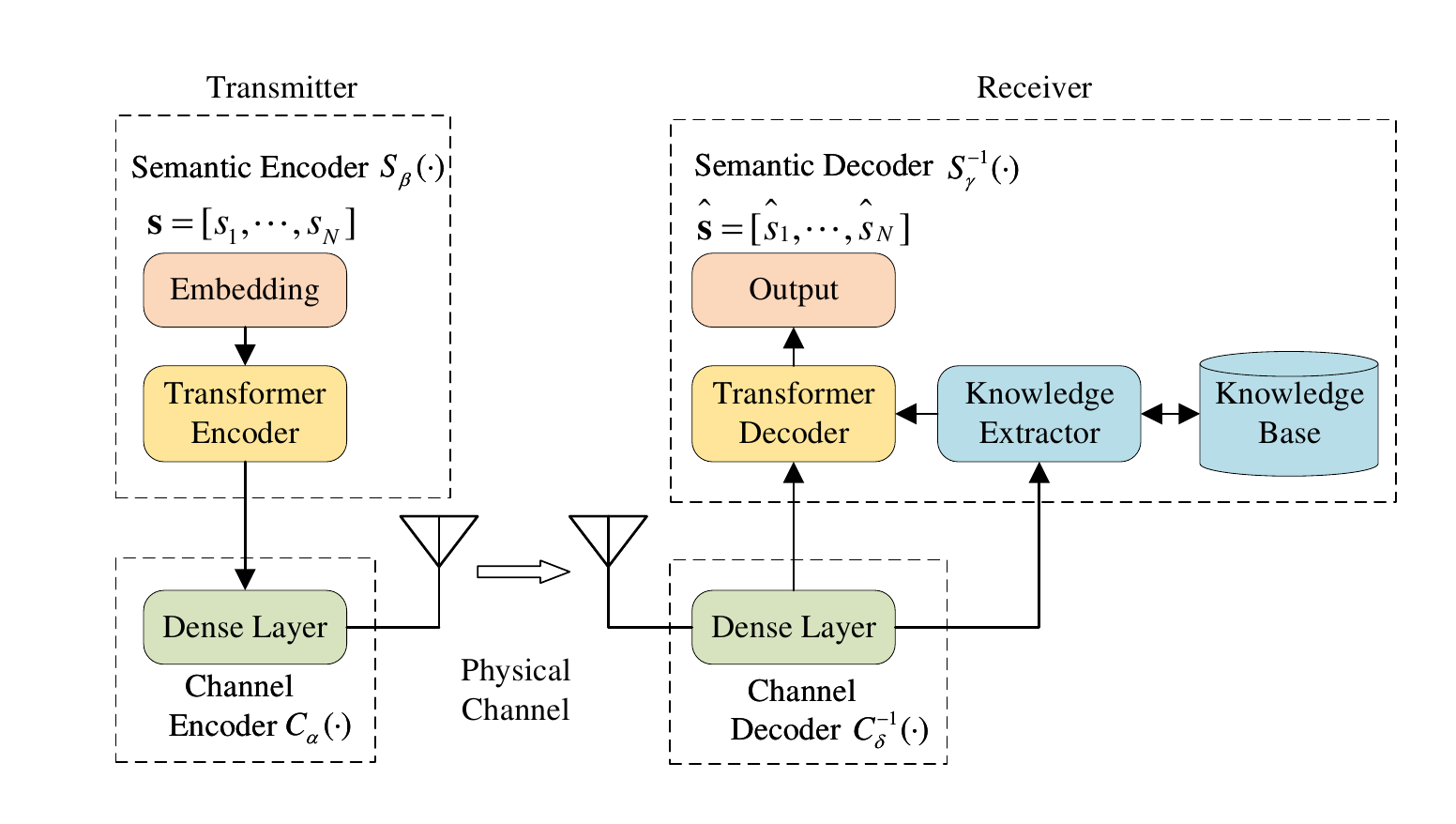}}
\caption{The framework of proposed knowledge graph enhanced semantic communication system.}
\label{fig1}
\end{figure}

\begin{table}[tbp]
\caption{Notations used in this paper}
\begin{center}
\scalebox{1}{
\begin{tabular}{ll}
\toprule
Notation & Definition \\ \midrule
$S_\beta(\cdot), S^{-1}_\gamma(\cdot)$ & Semantic encoder and decoder \\
$C_\alpha(\cdot), C^{-1}_\delta(\cdot)$ & Channel encoder and decoder \\
$K_\theta(\cdot)$ & Knowledge extractor \\
$\mathbf{s}, \hat{\mathbf{s}}$ & Input and decoded sentence \\ 
$\mathbf{h}$ & Semantically encoded vector \\
$\hat{\mathbf{h}}$ & Channel decoded vector \\
$\mathbf{x}$ & Transmitted signal  \\
$\mathbf{y}$ & Received signal  \\
$\mathbf{k}$ & Knowledge vector \\
$\mathbf{t}$ & Index vector of extracted triples \\
$n_t$ & Number of triples in the knowledge base \\
$N$ & Length of sentence \\
$w$ & Weight parameter for knowledge extraction\\
$f_{k}(\cdot)$ & Knowledge embedding process\\
\bottomrule
\end{tabular}
}
\label{notation_table}
\end{center}
\end{table}
A semantic communication system generally encompasses a semantic encoder and decoder, which can be depicted in Fig.~\ref{fig1}. Without loss of generality, we denote the input sentence $\mathbf{s} = [s_1, s_2,...,s_N] \in \mathbb{N}^{N}$, where $s_i$ represents the $i$-th word (i.e., token) in the sentence. In particular, the transmitter consists of two modules, that is, the semantic encoder and the channel encoder. The semantic encoder $S_\beta(\cdot)$ extracts the semantic information in the content and represents it as a vector $\mathbf{h} \in \mathbb{R}^{N \times d_s}$, where $d_s$ is the dimension of each semantic symbol. Mathematically, 
\begin{equation}
\mathbf{h} = S_\beta(\mathbf{s}), \label{semantic_encoding}
\end{equation}
and then the channel encoder $C_\alpha(\cdot)$ encodes $\mathbf{h}$ into symbols that can be transmitted over the physical channel as
\begin{equation}
\mathbf{x} = C_\alpha(\mathbf{h}), \label{channel_encoding}
\end{equation}
where $\mathbf{x} \in \mathbb{C}^{N \times c}$ is the channel vector for transmission, $c$ is the number of symbols for each token. 
Given that $\mathbf{y} \in \mathbb{C}^{N \times c}$ is the vector of received symbols after transmitting $\mathbf{x}$ over the physical channel, $\mathbf{y}$ can be formulated as
\begin{equation}
\mathbf{y} = \mathbf{Hx} + \mathbf{n}, \label{channel_transmitting}
\end{equation}
where $\mathbf{H}$ denotes the channel matrix and $\mathbf{n} \sim \mathcal{N}(0, \sigma^2\mathbf{I})$ is the additive white Gaussian noise (AWGN).

After receiving $\mathbf{y}$, the receiver first decodes the content from symbols with the channel decoder $C^{-1}_\delta(\cdot)$ and gets the decoded vector $\hat{\mathbf{h}} \in \mathbb{R}^{N \times d_s}$,
\begin{equation}
\hat{\mathbf{h}} = C^{-1}_\delta(\mathbf{y}). \label{channel_decoding}
\end{equation}

Notably, semantic communications implicitly rely on some prior knowledge between the transmitter and receiver for the joint training process. However, different from such prior knowledge, the knowledge base in our model refers to some factual triples and can be located at the receiver side. To exploit the knowledge base, a knowledge extractor is further applied at the receiver side to extract and integrate relevant knowledge from the received signal to yield the aggregated knowledge $\textbf{k}$. In particular, the knowledge extracting and aggregating process can be formulated as
\begin{equation}
\mathbf{k} = K_\theta(\hat{\mathbf{h}}), \label{knowledge_extraction}
\end{equation}
where $K_\theta(\cdot)$ represents the knowledge extractor. 

Then the knowledge enhanced semantic decoder $S^{-1}_\gamma(\cdot)$ leverages the channel decoded vector $\hat{\mathbf{h}}$ and the extracted knowledge vector $\mathbf{k}$ to obtain the received message $\hat{\mathbf{s}} = [\hat{s}_1, \hat{s}_2,...,\hat{s}_N]$
\begin{equation}
\hat{\mathbf{s}} = S^{-1}_\gamma(\hat{\mathbf{h}} \  || \  \mathbf{k}), \label{semantic_decoding}
\end{equation}
where $S^{-1}_\gamma(\cdot)$ stands for the knowledge enhanced semantic decoder, and $||$ indicates a concatenation operator. 

The accuracy of semantic communication is determined by the semantic similarity between the sent and received contents. In order to minimize the semantic errors between $\mathbf{s}$ and $\hat{\mathbf{s}}$, the loss function can take account of the cross entropy of the two vectors
\begin{equation}
\mathcal{L}_\text{model} = - \sum_{i=1}^{N}(q(s_i)\log{p(\hat{s}_i)}),
\label{semantic_similarity}
\end{equation}
where $\mathcal{L}_\text{model}$ is the loss function, $q(s_i)$ is the one-hot representation of $s_i \in \mathbf{s}$, and $p(\hat{s}_i)$ is the predicted probability of the $i$-th word. 

Instead of using traditional communication modules for physical-layer transmission, most existing studies have chosen to utilize end-to-end DNNs to accomplish the whole communication process, as shown in Fig.~\ref{fig1}. The semantic encoders and decoders are typically based on transformers \cite{vaswani2017attention}. Meanwhile, the channel encoding and decoding part can be viewed as an autoencoder implemented by fully connected layers. The whole semantic communication process is then reformulated as a sequence-to-sequence problem. Based on these DNNs, in this paper, we primarily focus on developing appropriate implementation of the knowledge extractor in \eqref{knowledge_extraction} and the knowledge-enhanced decoder in \eqref{semantic_decoding} to minimize the model loss function.

\section{A Knowledge Extractor Model based Semantic Decoder}

\subsection{The Design of the Knowledge Extractor}

\begin{figure}[t]
\centerline{\includegraphics[width=\linewidth]{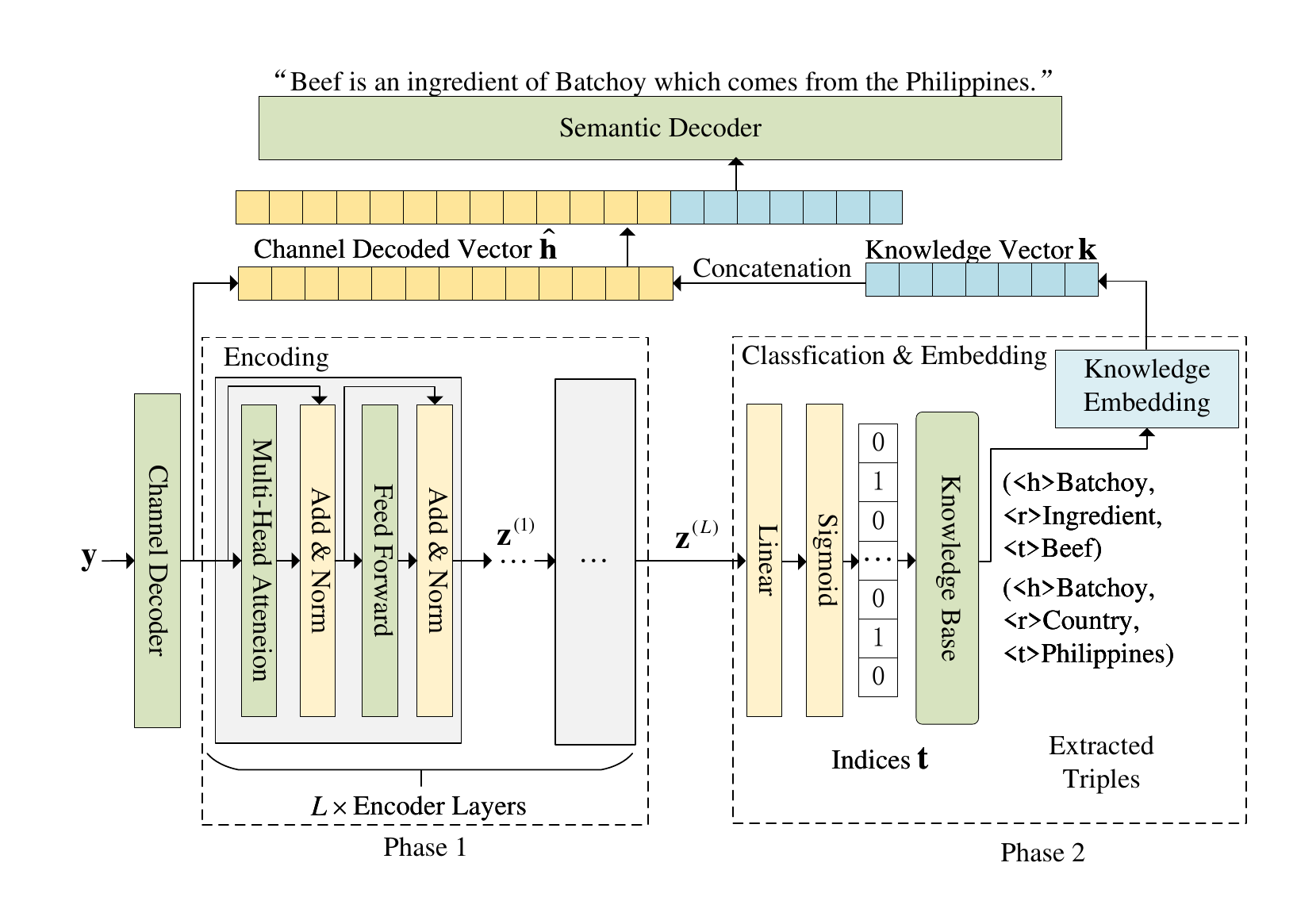}}
\caption{The knowledge graph enhanced semantic decoder.}
\label{fig2}
\end{figure}

In this part, we discuss the implementation of knowledge extractor enhanced semantic decoding task. The whole knowledge extraction process, as shown in Fig.~\ref{fig2}, can be divided into two phases. The first phase executes an embedding task to obtain a representation of the decoded vector with the transformer encoders. In the second phase, we try to find all corresponding triples of the representation with a multi-label classifier, and then embed the triples into a compressed format to assist the final decoding.

In particular, in order to extract the semantic representation, we adopt a model composed of a stack of $L$ identical transformer encoders, each of them consisting a multi-head attention mechanism, as well as some feed-forward and normalization sublayers \cite{vaswani2017attention}. In particular, without loss of generality, assuming that $\mathbf{z}^{(l-1)}$ is the output of the $({l-1})$-th encoder layer, where $\mathbf{z}^{(0)}$ is equivalent to $\hat{\mathbf{h}}$ for the input layer, the self-attention mechanism of the $l$-th layer could be represented as
\begin{equation}
\text{Attention}(\mathbf{z}^{(l-1)}) = \text{softmax}(\frac{\mathbf{Q}^{(l)}\mathbf{K}^{(l)\mathrm{T}}}{\sqrt{d_k}})\mathbf{V}^{(l)},
\label{attention}
\end{equation}
where $\mathbf{Q}^{(l)} = \mathbf{z}^{(l-1)} \mathbf{W}_Q^{(l)} $, $\mathbf{K}^{(l)} = \mathbf{z}^{(l-1)} \mathbf{W}_K^{(l)} $, $\mathbf{V}^{(l)} = \mathbf{z}^{(l-1)} \mathbf{W}_V^{(l)} $. $\mathbf{W}_Q^{(l)}$, $\mathbf{W}_K^{(l)}$ and $\mathbf{W}_V^{(l)}$ are the projection matrices of the $l$-th layer, and $d_k$ is the dimension of model. Furthermore, $\mathbf{z}^{(l-1)}$ is added to the calculated result $\text{Attention}(\mathbf{z}^{(l-1)})$ via a normalized residual connection, that is,
\begin{equation}
\mathbf{a}^{(l)} = \text{LayerNorm}(\text{Attention}(\mathbf{z}^{(l-1)}) + \mathbf{z}^{(l-1)}),
\label{add_and_norm}
\end{equation}
where $\mathbf{a}^{(l)}$ is the output, $\text{LayerNorm}(\cdot)$ denotes a layer normalization operation. Afterwards, a feed-forward network is involved as $\text{FFN}(\mathbf{a}^{(l)}) = \text{max}(0, \mathbf{a}^{(l)} \mathbf{W}_\text{F1}^{(l)} + \mathbf{b}_\text{F1}^{(l)}) \mathbf{W}_\text{F2}^{(l)} + \mathbf{b}_\text{F2}^{(l)}$, where $\mathbf{W}_\text{F1}^{(l)}$, $\mathbf{W}_\text{F2}^{(l)}$, $\mathbf{b}_\text{F1}^{(l)}$ and $\mathbf{b}_\text{F2}^{(l)}$ are parameters in the feed-forward layer of the $l$-th encoder block. Next, we adopt a residual connection and a layer normalization
\begin{equation}
\mathbf{z}^{(l)} = \text{LayerNorm}(\text{FFN}(\mathbf{a}^{(l)}) + \mathbf{a}^{(l)}).
\label{add_and_norm2}
\end{equation}

After $L$ layers of encoding, the embedding representation $\mathbf{z}^{(L)}$ of the channel decoded vector $\hat{\mathbf{h}}$ is obtained. Then a multi-label classifier is adopted to compute a indicator vector $\mathbf{t}$ of the triples associated with the representation
\begin{equation}
\mathbf{t} = \text{sigmoid}(\mathbf{z}^{(L)} \mathbf{W}_t  + \mathbf{b}_t),
\label{prediction}
\end{equation}
where $\mathbf{t} = [\hat{t}_1, \cdots, \hat{t}_{n_t}] \in \mathbb{R}^{n_t}, \hat{t}_i \in [0, 1]$ for all $i = 1\cdots n_t$. $n_t$ denotes the number of triples in the knowledge base, $\mathbf{W}_t$ and $\mathbf{b}_t$ are parameters of the classifier. If $\hat{t}_i \ge 0.5$, the triple $m_i$ corresponding to index $i$ is predicted to be relevant to the received content. 

Ultimately, the obtained relevant factual triples $\{m_i\}$ predicted by the model are embedded into a vector $\mathbf{k} = f_{k}(\{m_i\})$, where the embedding process is abstractly represented as $f_{k}(\cdot)$. In particular, rather than compute the embedding of the entity and relationship separately, we choose to integrate the triples into one compressed format. Subsequently, as in \eqref{semantic_decoding}, the knowledge vector is concatenated with the decoding vector and fed into the semantic decoder.

\subsection{The Training Methodology}

In order to train the knowledge extractor $K_\theta(\cdot)$, a complete semantic communication model is first required. During the training, the sentences are first sent through the transmitter via the channel and then the receiver decoded vector is fed into the knowledge extractor. Afterwards, the knowledge extractor is trained by gradient descent with the frozen parameters of the transmitter. Since the number of negative labels are much more than that of the positive labels in the classification, the weighted Binary Cross Entropy (BCE) is utilized as the loss function, which could be represented as
\begin{equation}
\mathcal{L}_\text{knowledge} = \sum_{i=1}^{n_t} -w_i [t_i \cdot \log{\hat{t}_i} + (1 - t_i) \cdot \log(1 - \hat{t}_i)], \label{loss_function}
\end{equation}
where $t_i \in \{0, 1\}$ represents the training label, and $\hat{t}_i$ is the prediction output in \eqref{prediction}. $w_i$ is the weight of $i$-th position, related to the hyperparameter $w$. For $t_i = 0$, $w_i = w$; otherwise $w_i = 1 - w$. Increasing $w$ can result in a more sensitive extractor, but it also brings an increase in the false positive rate.

The training complexity of a knowledge extractor is $O(LN^2 \cdot d_k)$, which is the same order as the transformer encoder. Notably, the knowledge extractor is not limited to the conventional transformer structure, but can also be applied to different transformer variants, such as Universal Transformer (UT)\cite{zhou2021semantic}. With the self-attention mechanism, the extracted factual triples can provide additional prior knowledge to the semantic decoder and therefore improve the performance of the decoder. Typically, the knowledge vector is concatenated to the received message, rather than being merged into the source signal as previous works suggested. This architecture ensures that when the extractor is of little avail, it can still function as a standard encoder-decoder transformer structure, while avoiding possible semantic losses introduced by the knowledge extraction procedure. Therefore, even if the knowledge extractor fails to find any relevant knowledge, the model still performs comparably to the baseline.
\begin{algorithm}[tbp]
	\caption{The Semantic Communication Process with Knowledge Graph Enhanced Receiver} 
	\label{algorithm1}
	\begin{algorithmic}[1]
	\STATE \textbf{Require}: models $S_\beta(\cdot)$, $C_\alpha(\cdot)$, $S^{-1}_\gamma(\cdot)$, $C^{-1}_\delta(\cdot)$ and $K_\theta(\cdot)$ \\
	\STATE \textbf{Input}: tokenized sentence $\mathbf{s}$ \\
    \STATE \textbf{Output}: decoded sentence $\hat{\mathbf{s}}$
    \STATE \textbf{Transmitter}:
	\STATE \quad Semantic encoding: $\mathbf{h} \gets (S_\beta(\mathbf{s}))$
	\STATE \quad Channel encoding: $\mathbf{x} \gets (C_\alpha(\mathbf{h}))$
	\STATE \quad Transmit $\mathbf{x}$ over the physical channel: $\mathbf{y} \gets \mathbf{Hx} + \mathbf{n}$
	\STATE \textbf{Receiver}: 
	\STATE \quad Channel decoding: $\hat{\mathbf{h}} \gets C^{-1}_\delta(\mathbf{y})$
	\STATE \quad Knowledge extraction $K_\theta(\cdot)$:
	\STATE \quad \quad Compute the embedding representation $\mathbf{z}^{(L)}$
	\STATE \quad \quad $\mathbf{t} \gets \text{sigmoid}(\mathbf{z}^{(L)} \mathbf{W}_t  + \mathbf{b}_t)$
	\STATE \quad \quad Find the triples $\{m_i\}$ where $\hat{t}_i \ge 0.5$
 	\STATE \quad \quad Knowledge embedding: $\mathbf{k} \gets f_{k}(\{m_i\})$
    \STATE \quad Semantic decoding: $\hat{\mathbf{s}} \gets S^{-1}_\gamma(\hat{\mathbf{h}} \  || \  \mathbf{k})$
	\end{algorithmic} 
\end{algorithm}

\parskip=0pt
\section{Numerical Results}
\subsection{Dataset and Parameter Settings}

The dataset used in the numerical experiment is based on WebNLG v3.0\cite{gardent2017creating}, which consists of data-text pairs where the data is a set of triples extracted from DBpedia and the text is the verbalization of these triples. In this numerical experiment, the weight parameter $w$ is set to 0.02, while the learning rate is set to $10^{-4}$. Moreover, we set the dimension of the dense layer as $128 \times 16$, and adopt an 8-head attention in transformer layer. The detailed settings of the proposed system are shown in Table~\ref{tab2}. We train the models based on both the conventional transformer and UT \cite{zhou2021semantic}. Besides, we adopt two metrics to evaluate their performance, that is, 1-gram Bilingual Evaluation Understudy (BLEU)\cite{papineni2002bleu} score for measuring word-level accuracy and Sentence-Bert\cite{reimers2019sentence} score for measuring semantic similarity. Notably, Sentence-Bert is a Siamese Bert-network model that generates fixed-length vector representations for sentences, while the Sentence-Bert score computes the cosine similarity of embedded vectors.

\begin{figure*}[htbp]
\vspace{-0.5cm}
\begin{minipage}[t]{0.32\textwidth}
\centering
\includegraphics[width=\textwidth]{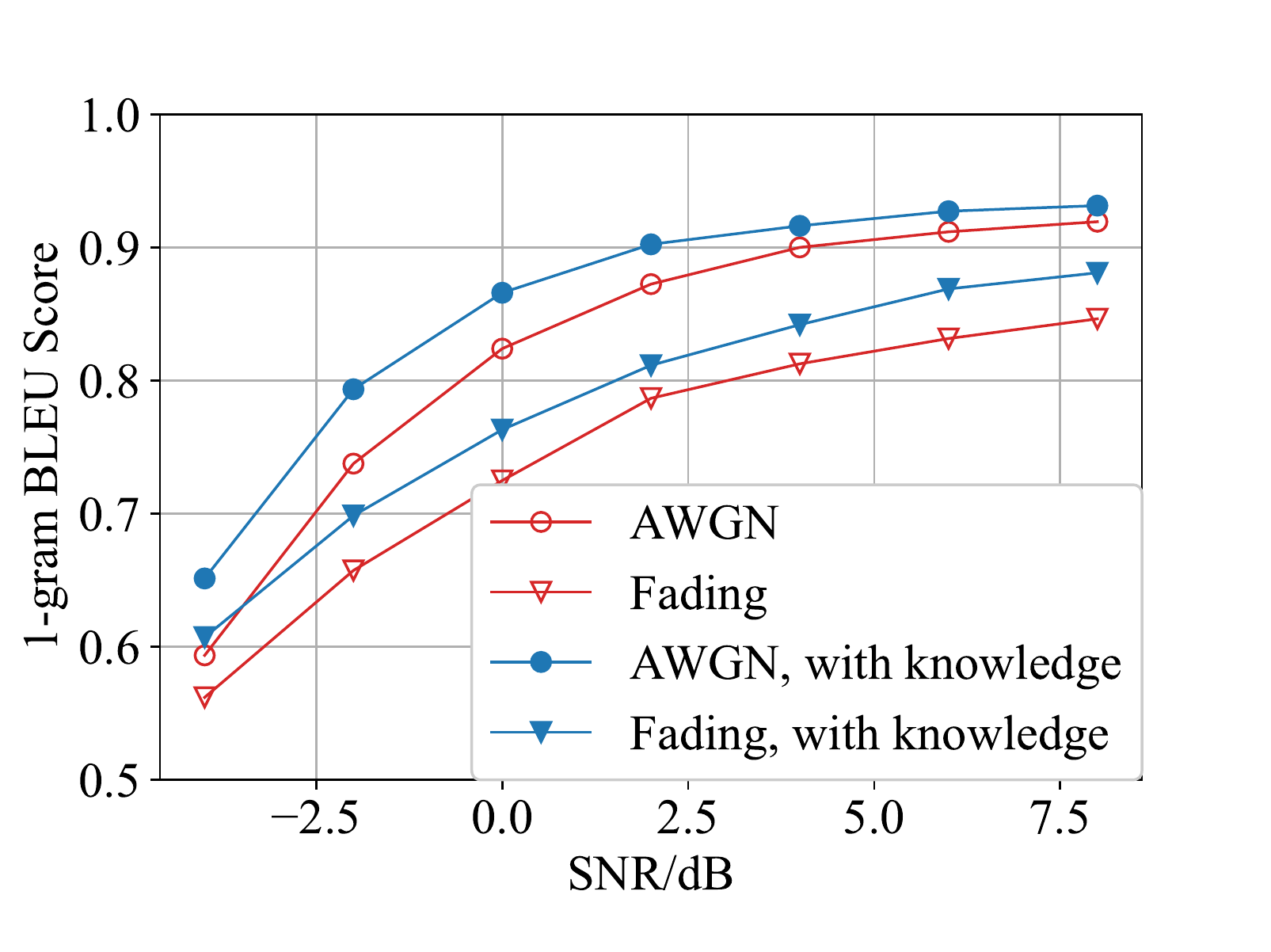}
\caption{The transformer model performance\\of BLEU score with respect to SNR.}
\label{fig3}
\end{minipage}
\begin{minipage}[t]{0.32\textwidth}
\centering
\includegraphics[width=\textwidth]{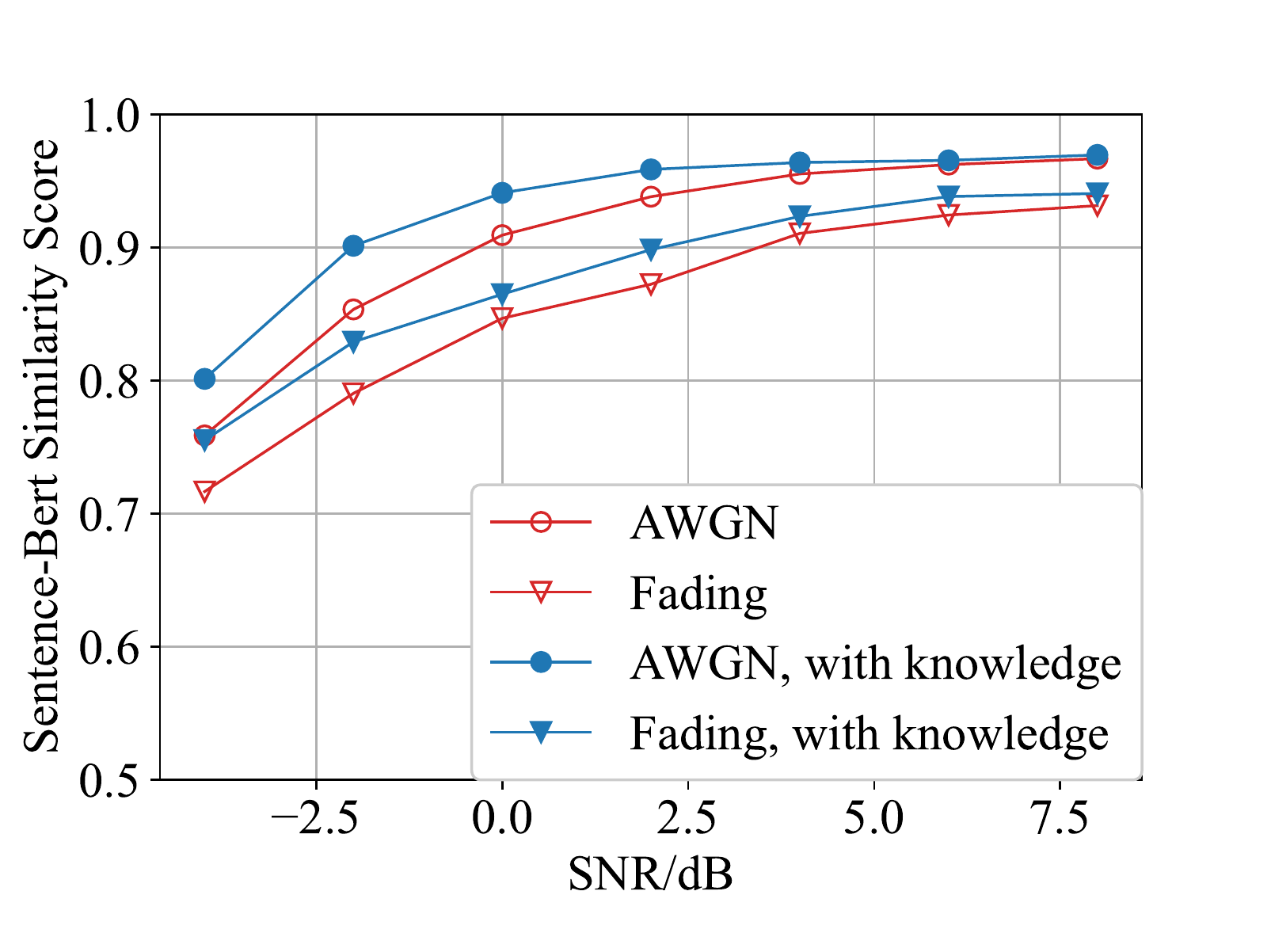}
\caption{The transformer model performance\\of Sentence-Bert score with respect to SNR.}
\label{fig4}
\end{minipage}
\begin{minipage}[t]{0.32\textwidth}
\centering
\includegraphics[width=\textwidth]{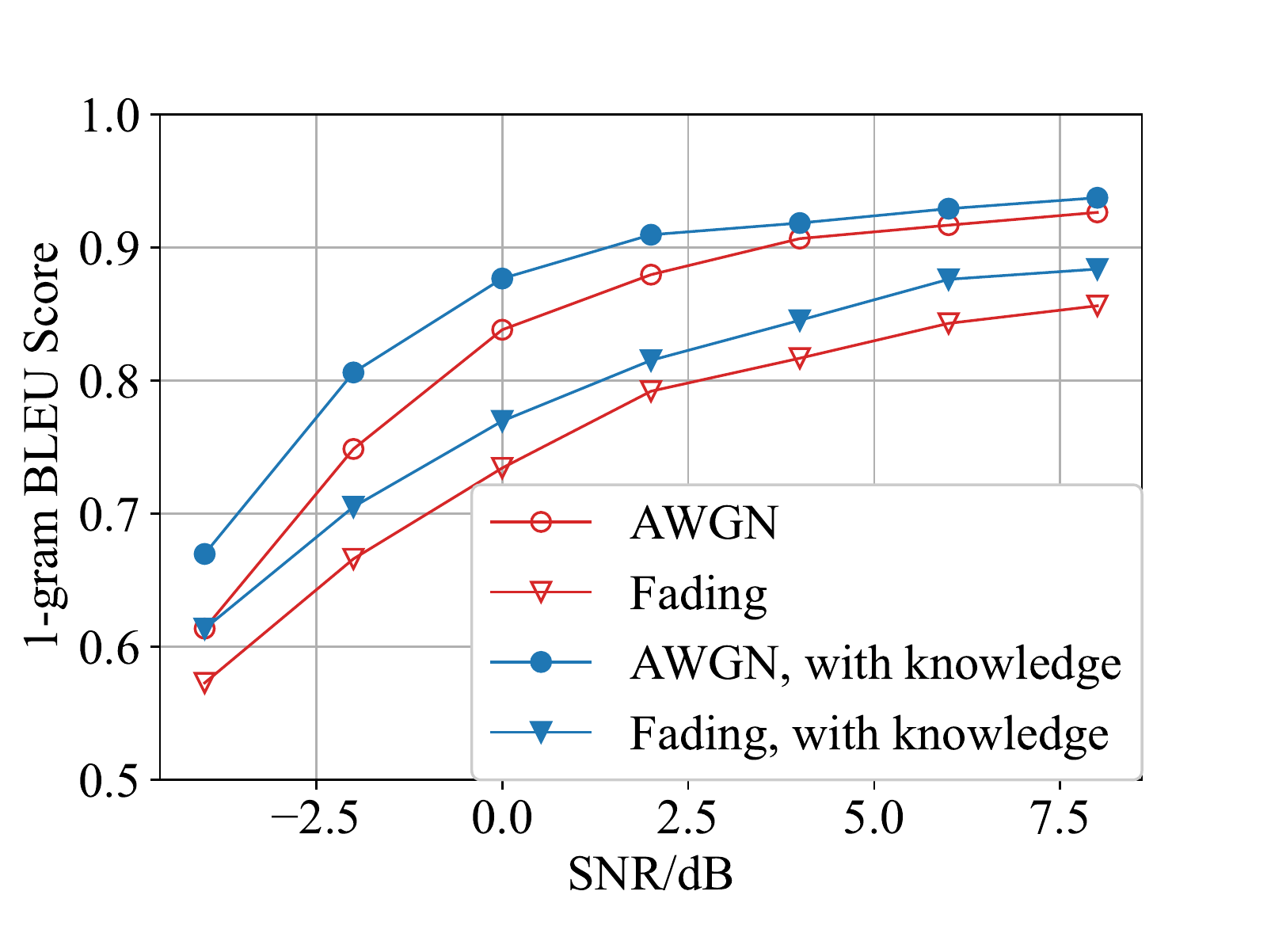}
\caption{The Universal Transformer model\\performance of BLEU score with respect to SNR.}
\label{fig5}
\end{minipage}
\end{figure*}

\subsection{Numerical Results}

\begin{figure*}[htbp]
\vspace{-0.4cm}
\begin{minipage}[t]{0.32\textwidth}
\centering
\includegraphics[width=\textwidth]{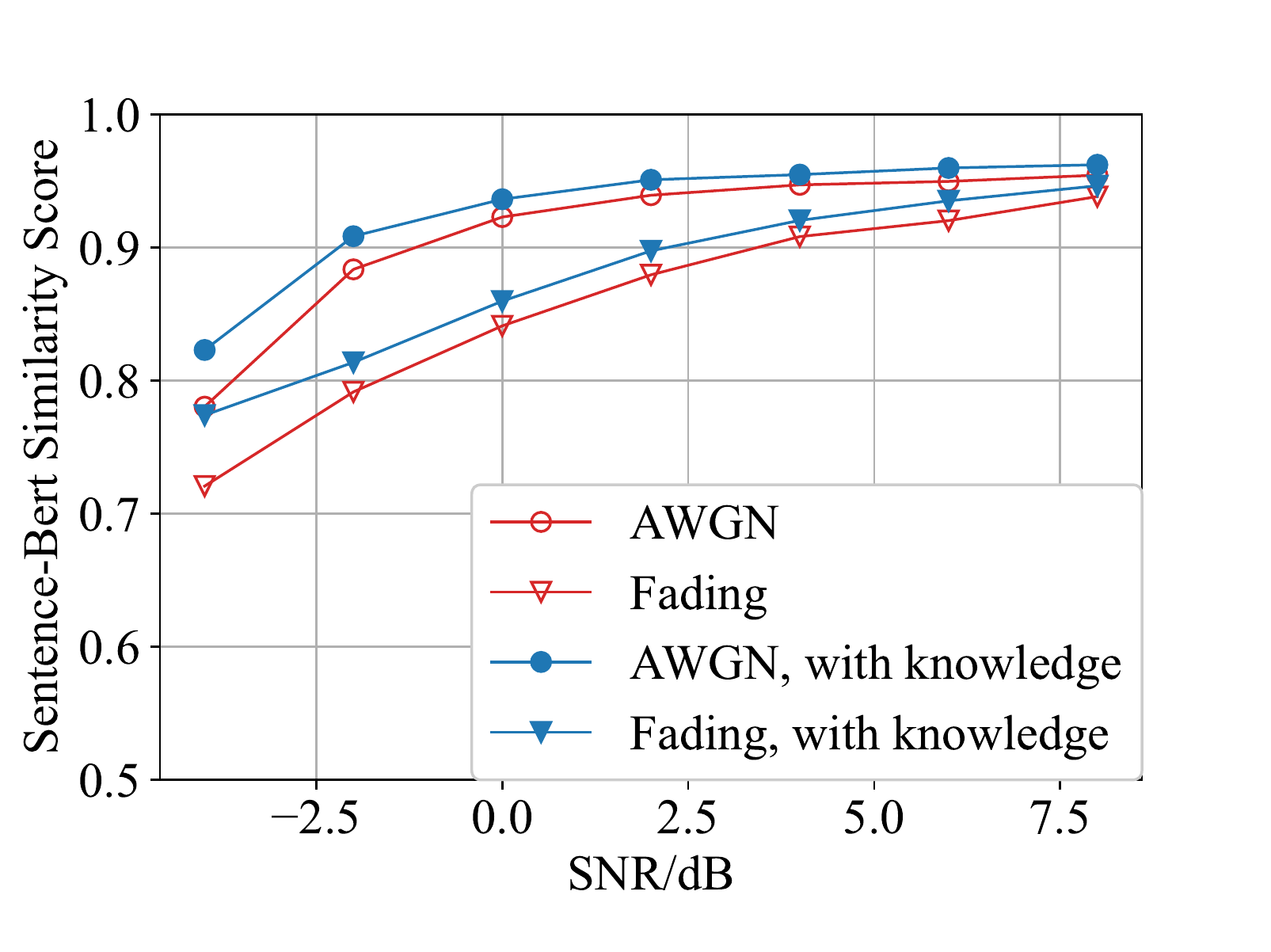}
\caption{The Universal Transformer model\\performance of Sentence-Bert score with respect\\to SNR.}
\label{fig6}
\end{minipage}
\begin{minipage}[t]{0.32\textwidth}
\centering
\includegraphics[width=\textwidth]{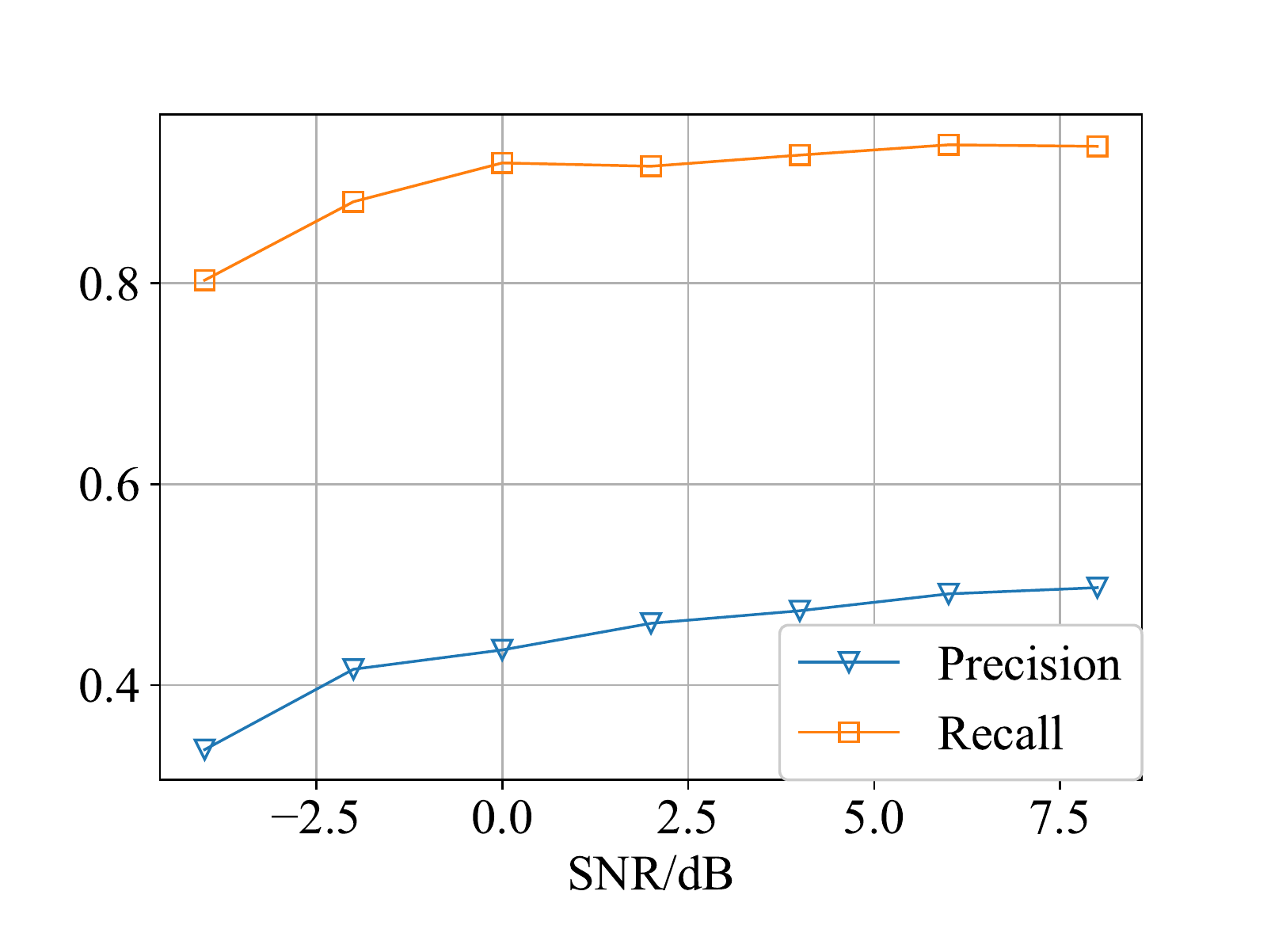}
\caption{The performance of knowledge extractor.}
\label{fig7}
\end{minipage}
\begin{minipage}[t]{0.32\textwidth}
\centering
\includegraphics[width=\textwidth]{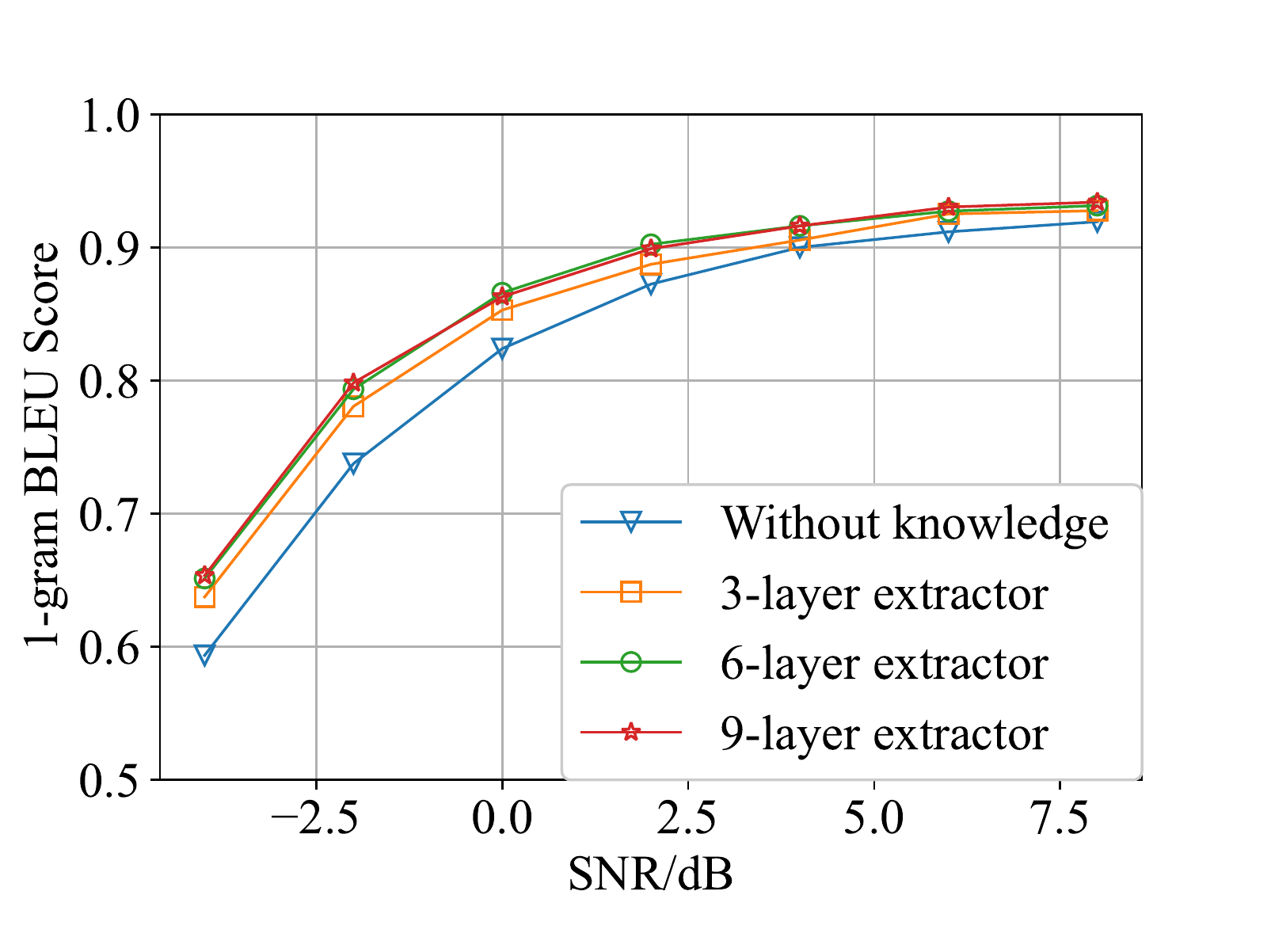}
\caption{The performance comparison under different numbers of encoder
layers for extractor.}
\label{fig8}
\end{minipage}
\end{figure*}

Fig.~\ref{fig3} and Fig.~\ref{fig4} show the BLEU score and Sentence-Bert score of the transformer model with respect to the signal-to-noise ratio (SNR), respectively. It can be observed that the assistance of the knowledge extractor could significantly contribute to improving the performance. In particular, regardless of the channel type, the knowledge extractor can always bring more than 5\% improvement in BLEU under low SNR scenarios. For the Sentence-Bert score, the knowledge-enhanced receiver also shows a similar performance improvement. This demonstrates that the proposed scheme can improve the comprehension of semantics at the receiver side. Fig.~\ref{fig5} and Fig.~\ref{fig6} demonstrate the performance of the UT model under both the BLEU and Sentence-Bert metric, and a similar performance improvement could also be observed.

\begin{table}[tbp]
\caption{Experimental Settings}
\begin{center}
\scalebox{1.0}{
\begin{tabular}{ll}
\toprule
Parameter & value \\
\midrule
Train dataset size & 24,467 \\ 
Test dataset size  & 2,734  \\
Weight parameter $w$  & 0.02  \\
DNN Optimizer & Adam \\
Batch size & 32 \\
Model dimension & 128 \\
Learning rate & $10^{-4}$ \\
Channel vector dimension & 16 \\
The number of multi-heads & 8 \\
\bottomrule
\end{tabular}
}
\label{tab2}
\end{center}
\end{table}

\begin{figure}[htbp]
\vspace{-0.4cm}
\centerline{\includegraphics[width=0.62\columnwidth]{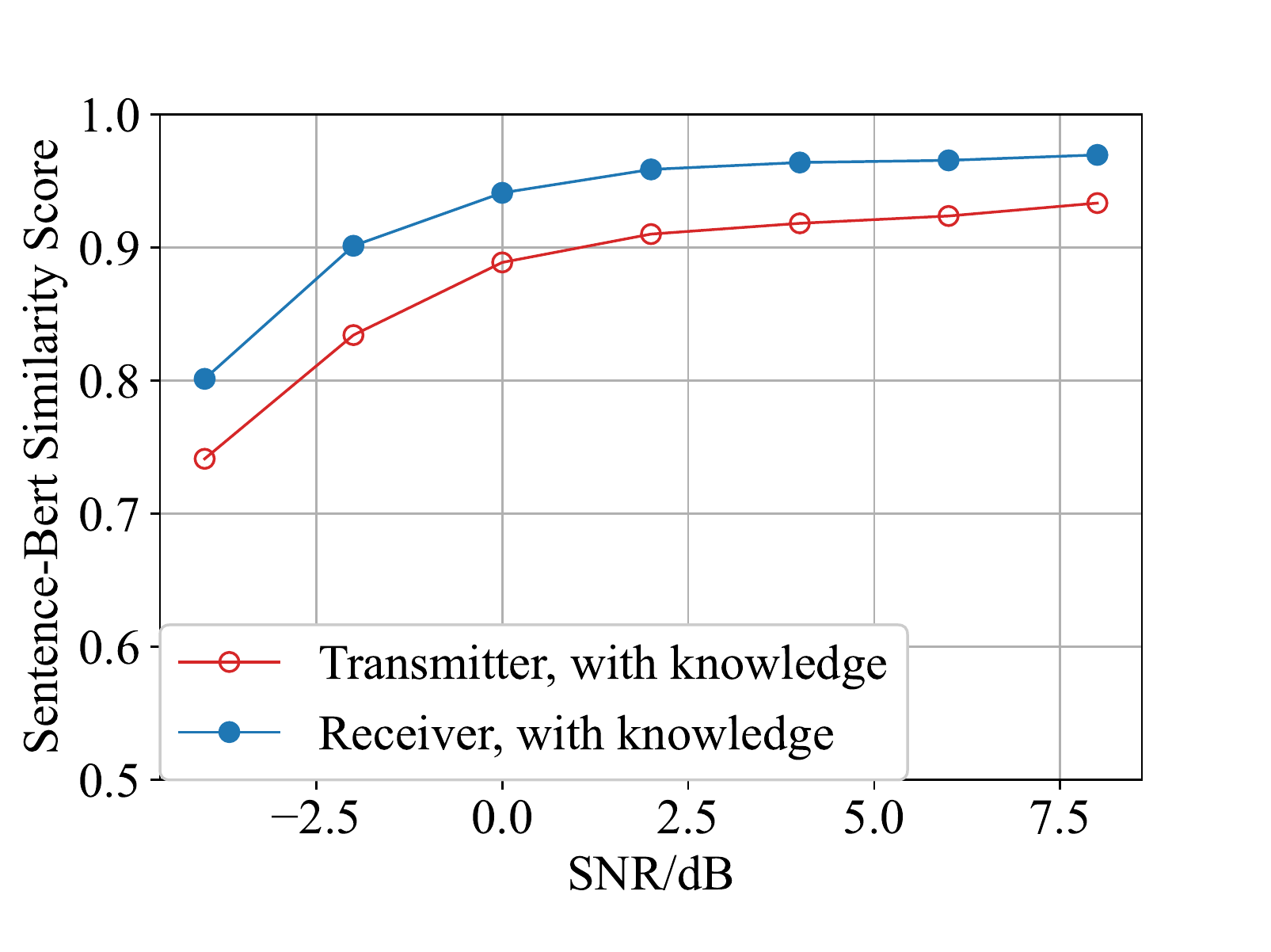}}
\caption{The performance comparison of the proposed scheme with a transmitter-based scheme.}
\label{fig9}
\end{figure}

On the other hand, we test the performance of the knowledge extractor under different SNRs. As shown in Fig.~\ref{fig7}, the extractor model can obtain a recall rate of over 90\%. However, the received content may be polluted by noise, resulting in an increase in false positives and leading to a large gap between precision and recall. The number of encoder layers in the knowledge extractor may also affect the performance of the model. Therefore, we also implement the knowledge extractor with different number of transformer encoder layers and present the performance comparison in Fig.~\ref{fig8}. It can be observed that the 6-layer model performs slightly better than the 3-layer model. However, the performance remains almost unchanged when it further to 9 layers. Furthermore, in addition to utilize a fixed model trained at certain SNR, it is also possible to leverage several SNR-specific models, each corresponding to a specific SNR. Table~\ref{tab3} demonstrates the performance comparison between 0dB-specific and fixed model. It can be observed that compared to the fixed model, the SNR-specific model could yield superior performance improvements. As a comparison, we also implement a scheme that utilize knowledge extractor for semantic encoding at the transmitter. Fig.~\ref{fig9} presents the corresponding simulation results, and it can be observed that this transmitter-based scheme is significantly inferior than the proposed scheme.

\begin{table}[tbp]
\centering
\vspace{-0.4cm}
\caption{The performance comparison between a fixed extractor model and SNR-specific models.}
\scalebox{0.95}{
\begin{tabular}{ccc}
\toprule
SNR/dB & Fixed & SNR-specific  \\ \midrule
-4 &  0.6514 &  0.6718 \\ %\midrule
-2 &  0.7936 &  0.8126 \\ %\midrule
0  &  0.8661 &  0.8661 \\ %\midrule
2  &  0.9025 &  0.9134 \\ %\midrule
4  &  0.9164 &  0.9201 \\
\bottomrule
\end{tabular}
}
\label{tab3}
\end{table}

\section{Conclusion}
In this paper, we have proposed a knowledge graph enhanced semantic communication framework in which the receiver can utilize prior knowledge from the knowledge graph for semantic decoding while requiring no additional modifications to the transmitter architecture. Specifically, we have designed a knowledge extractor to find the factual triples associated with the received noisy sentences. Simulation results on the WebNLG dataset have shown that our proposed system is able to exploit the prior knowledge in the knowledge base much more deeply and obtain performance gains at the receiver side. In the future, we will investigate the joint optimization of both the transmitter and the receiver enhanced by the knowledge graph.

\bibliographystyle{IEEEtran}
\bibliography{reference}

\begin{thebibliography}{10}
\providecommand{\url}[1]{#1}
\csname url@samestyle\endcsname
\providecommand{\newblock}{\relax}
\providecommand{\bibinfo}[2]{#2}
\providecommand{\BIBentrySTDinterwordspacing}{\spaceskip=0pt\relax}
\providecommand{\BIBentryALTinterwordstretchfactor}{4}
\providecommand{\BIBentryALTinterwordspacing}{\spaceskip=\fontdimen2\font plus
\BIBentryALTinterwordstretchfactor\fontdimen3\font minus
  \fontdimen4\font\relax}
\providecommand{\BIBforeignlanguage}[2]{{%
\expandafter\ifx\csname l@#1\endcsname\relax
\typeout{** WARNING: IEEEtran.bst: No hyphenation pattern has been}%
\typeout{** loaded for the language `#1'. Using the pattern for}%
\typeout{** the default language instead.}%
\else
\language=\csname l@#1\endcsname
\fi
#2}}
\providecommand{\BIBdecl}{\relax}
\BIBdecl

\bibitem{xie2021deep}
H.~Xie, Z.~Qin \emph{et~al.}, ``Deep learning enabled semantic communication
  systems,'' \emph{IEEE Transactions on Signal Processing}, vol.~69, pp.
  2663--2675, 2021.

\bibitem{zhou2021semantic}
Q.~Zhou, R.~Li \emph{et~al.}, ``Semantic communication with adaptive universal
  transformer,'' \emph{IEEE Wireless Commun. Lett.}, vol.~11, no.~3, pp.
  453--457, 2021.

\bibitem{jiang2022deep}
P.~Jiang, C.-K. Wen \emph{et~al.}, ``Deep source-channel coding for sentence
  semantic transmission with {HARQ},'' \emph{IEEE Trans. Commun.}, vol.~70,
  no.~8, pp. 5225--5240, 2022.

\bibitem{zhou2022adaptive}
Q.~Zhou, R.~Li \emph{et~al.}, ``Adaptive bit rate control in semantic
  communication with incremental knowledge-based {HARQ},'' \emph{IEEE Open J.
  Commun. Soc.}, vol.~3, pp. 1076--1089, 2022.

\bibitem{hu2022robust}
Q.~Hu, G.~Zhang \emph{et~al.}, ``Robust semantic communications against
  semantic noise,'' in \emph{Proc. VTC2022-FALL}, 2022.

\bibitem{ji2021survey}
S.~Ji, S.~Pan \emph{et~al.}, ``A survey on knowledge graphs: Representation,
  acquisition, and applications,'' \emph{IEEE Trans. Neural Netw. Learn.
  Syst.}, vol.~33, no.~2, pp. 494--514, 2022.

\bibitem{wang2021performance}
Y.~Wang, M.~Chen \emph{et~al.}, ``Performance optimization for semantic
  communications: An attention-based learning approach,'' in \emph{Proc. IEEE
  Global Commun. Conf. (GLOBECOM)}, Madrid, Spain, Dec. 2021.

\bibitem{zhou2022cognitive}
F.~Zhou, Y.~Li \emph{et~al.}, ``Cognitive semantic communication systems driven
  by knowledge graph,'' in \emph{Proc. ICC}, Seoul, South Korea, May 2022.

\bibitem{liang2022life}
J.~Liang, Y.~Xiao \emph{et~al.}, ``Life-long learning for reasoning-based
  semantic communication,'' in \emph{Proc. ICC Workshops}, Seoul, South Korea,
  May 2022.

\bibitem{jiang2022reliable}
S.~Jiang, Y.~Liu \emph{et~al.}, ``Reliable semantic communication system
  enabled by knowledge graph,'' \emph{Entropy}, vol.~24, no.~6, p. 846, 2022.

\bibitem{choi2022unified}
J.~Choi, S.~W. Loke, and J.~Park, ``A unified approach to semantic information
  and communication based on probabilistic logic,'' \emph{IEEE Access},
  vol.~10, pp. 129\,806--129\,822, 2022.

\bibitem{muppavarapu2021knowledge}
V.~Muppavarapu, G.~Ramesh \emph{et~al.}, ``Knowledge extraction using semantic
  similarity of concepts from web of things knowledge bases,'' \emph{Data \&
  Knowledge Engineering}, vol. 135, p. 101923, 2021.

\bibitem{vaswani2017attention}
A.~Vaswani \emph{et~al.}, ``Attention is all you need,'' in \emph{Proc.
  NeurIPS}, Long Beach, USA, Dec. 2017.

\bibitem{gardent2017creating}
C.~Gardent, A.~Shimorina \emph{et~al.}, ``Creating training corpora for nlg
  micro-planning,'' in \emph{Proc. ACL}, Vancouver, Canada, Jul./Aug. 2017.

\bibitem{papineni2002bleu}
K.~Papineni, S.~Roukos \emph{et~al.}, ``Bleu: {A} method for automatic
  evaluation of machine translation,'' in \emph{Proceedings of the 40th annual
  meeting of the Association for Computational Linguistics}, 2002.

\bibitem{reimers2019sentence}
N.~Reimers and I.~Gurevych, ``Sentence-bert: Sentence embeddings using siamese
  bert-networks,'' in \emph{Proc. EMNLP-IJCNLP}, Hong Kong, China, 2019.

\end{thebibliography}

\end{document}